%% file: iclr2026_conference.tex
\documentclass{article} 
\usepackage{iclr2026_conference,times}

\input{math_commands.tex}

\definecolor{warningcolor}{RGB}{255, 0, 0}
\usepackage{hyperref}
\usepackage{url}
\usepackage{times}  
\usepackage{helvet}  
\usepackage{courier}  

\usepackage{wrapfig}
\usepackage{multirow}
\usepackage{color}
\usepackage{colortbl}
\usepackage{natbib}  
\usepackage{caption} 
\usepackage{amsmath}
\usepackage{amssymb}
\usepackage{mathtools}
\usepackage{booktabs}
\usepackage{multirow}
\usepackage{xspace}
\usepackage{siunitx}

\title{AI-Generated Text is Non-Stationary: Detection via Temporal Tomography}

\author{DeepScientist, \\
\textbf{Yixuan Weng}, \textbf{Minjun Zhu}, \textbf{Luodan Zhang}, \textbf{Zhen Lin}, \textbf{Guangsheng Bao}, \textbf{Yue Zhang} \\
School of Engineering, Westlake University\\
zhangyue@westlake.edu.cn}

\iclrfinalcopy 
\begin{document}

\maketitle

\begin{abstract}
The field of AI-generated text detection has evolved from supervised classification to zero-shot statistical analysis. However, current approaches share a fundamental limitation: they aggregate token-level measurements into scalar scores, discarding positional information about where anomalies occur. Our empirical analysis reveals that AI-generated text exhibits significant non-stationarity—statistical properties vary by 73.8\% more between text segments compared to human writing. This discovery explains why existing detectors fail against localized adversarial perturbations that exploit this overlooked characteristic. We introduce Temporal Discrepancy Tomography (TDT), a novel detection paradigm that preserves positional information by reformulating detection as a signal processing task. TDT treats token-level discrepancies as a time-series signal and applies Continuous Wavelet Transform to generate a two-dimensional time-scale representation, capturing both the location and linguistic scale of statistical anomalies. On the RAID benchmark, TDT achieves 0.855 AUROC (7.1\% improvement over the best baseline). More importantly, TDT demonstrates robust performance on adversarial tasks, with 14.1\% AUROC improvement on HART Level 2 paraphrasing attacks. Despite its sophisticated analysis, TDT maintains practical efficiency with only 13\% computational overhead. Our work establishes non-stationarity as a fundamental characteristic of AI-generated text and demonstrates that preserving temporal dynamics is essential for robust detection. Ours code are released at \url{https://github.com/ResearAI/TDT-Text-Detect}.

{\color{warningcolor} \normalsize WARMING:
We hereby declare that ours DeepScientist system performed approximately 95\% of the work presented in this paper. This includes the initial ideation, the design and execution of comparative experiments, the analysis of results, the literature review, the composition of the manuscript, the creation of the main figures, and the organization of the accompanying code repository. The role of the human authors was to supervise the AI’s operations. While we have diligently worked to minimize AI hallucinations and ensure the validity of the experimental results, we cannot fully guarantee against potential unintended outputs, system failures, or misleading conclusions. We therefore advise readers to approach this work with caution and to critically evaluate its findings before application.}
\end{abstract}

\newpage
\section{Introduction}

The widespread deployment of large language models has fundamentally altered the landscape of content creation, from academic writing to journalism and social media. This transformation brings unprecedented challenges for maintaining information integrity, as distinguishing between human and machine-generated text becomes increasingly difficult yet critically important \citep{Jawahar2020AutomaticDO}. The sophistication of modern language models enables not only wholesale generation of convincing text but also subtle modifications that preserve human-like qualities while introducing machine artifacts \citep{Su2025HACoDetAS, Zhang2024LLMasaCoauthorCM}.

Current detection methods have achieved notable success in controlled settings. Supervised approaches leverage large labeled datasets to learn discriminative features \citep{Solaiman2019ReleaseSD}, while zero-shot methods like DetectGPT exploit statistical properties inherent in model-generated text without requiring training data \citep{Mitchell2023DetectGPTZM}. Recent advances such as FastDetectGPT have further improved efficiency through conditional probability analysis \citep{Bao2023FastDetectGPTEZ}. However, these methods exhibit systematic failures when confronted with adversarial perturbations or domain shifts, suggesting fundamental limitations in their underlying assumptions. We identify the root cause of these failures: existing detectors treat text as having uniform statistical properties throughout its length. Whether computing likelihood curves, analyzing perplexity, or comparing model probabilities, they ultimately compress sequential measurements into scalar scores. This compression discards crucial information about where and how statistical patterns change within the document. Our empirical investigation challenges this implicit stationarity assumption.

Through systematic analysis of 200 documents using sliding window statistics, (details in Figure \ref{fig:analysis_results}a), we discover that AI-generated text exhibits fundamentally different temporal characteristics than human writing. Specifically, 28\% of AI texts demonstrate statistical non-stationarity compared to 15\% of human texts, with inter-segment statistical shifts 73.8\% larger in machine-generated content. This non-stationarity emerges from the autoregressive nature of language models—each token is generated based solely on preceding context, without the global planning and thematic coherence that characterize human writing. This finding has profound implications for detection robustness. Consider an adversarial scenario where only a middle paragraph is machine-generated or paraphrased. Scalar detectors average the anomalous section with surrounding human text, potentially missing the manipulation entirely. Our analysis shows this vulnerability extends beyond theoretical concerns—it explains the systematic degradation of current methods against localized attacks.

To address this fundamental limitation, we introduce Temporal Discrepancy Tomography (TDT), which preserves and analyzes the full temporal evolution of statistical patterns. Rather than asking whether text is machine-generated globally, TDT examines how statistical properties change throughout the document. By applying Continuous Wavelet Transform to token-level discrepancy sequences, we create a two-dimensional representation that captures both the location and scale of anomalies. The wavelet transform is particularly suited for this task as it excels at analyzing non-stationary signals, providing optimal time-frequency localization \citep{Daubechies1992TenLO}. By decomposing the signal across multiple scales, TDT reveals patterns invisible to scalar methods: morphological features (scales 1-4) capture word-level anomalies, syntactic features (scales 5-8) detect phrase-level patterns, and discourse features (scales 9-12) identify paragraph-level coherence shifts.

Extensive evaluation validates our approach. TDT achieves 0.855 AUROC on the RAID benchmark (7.1\% improvement) and excels on adversarial tasks with 14.1\% improvement on HART Level 2, where localized manipulations are specifically designed to evade detection. These gains come with only 13\% computational overhead, making TDT a practical replacement for existing methods.

Our contributions are threefold:
\begin{itemize}
\item We provide empirical evidence that non-stationarity is a fundamental characteristic of AI-generated text, not captured by current detection methods.
\item We demonstrate that preserving positional information through signal processing techniques significantly improves robustness, particularly against adversarial attacks.
\item We establish a new detection paradigm that analyzes temporal dynamics, achieving state-of-the-art performance while maintaining efficiency.
\end{itemize}

\section{Related Work}

The field of zero-shot AI text detection is largely built upon the foundational paradigm of analyzing log-probability discrepancies from a source language model. Seminal work like DetectGPT first hypothesized that machine text resides in areas of negative log-probability curvature, establishing a principle that inspired numerous follow-on methods \citep{Mitchell2023DetectGPTZM}. Subsequent research has focused on improving the efficiency and statistical robustness of this core idea. For instance, FastDetectGPT introduced sampling-based approximations to reduce computational overhead \citep{Bao2023FastDetectGPTEZ}, while other approaches like Binoculars leveraged the perplexity differences between two separate models to create a discriminative signal \citep{Hans2024SpottingLW}. Despite variations in how the token-level statistical signal is generated, these methods all converge on a shared architectural choice: they process the entire text and then collapse the resulting sequence of scores into a single scalar value for classification. Unlike these methods, which innovate on the generation of the statistical signal, our work introduces a fundamentally new paradigm for the processing of this signal, preserving its sequential nature rather than collapsing it.

Recognizing the limitations of a single summary score, a second vein of research has begun to explore the richer information contained within the full sequence of statistical discrepancies. T-Detect \citep{west2025tdetecttailawarestatisticalnormalization}, for example, addressed the heavy-tailed nature of log-probability distributions by applying a more robust Student's t-distribution normalization at the token level. More recently, \citet{xu-etal-2024-detecting} proposed moving from absolute likelihood values to relative ones and extracting features from the spectrum-view of the likelihood sequence, connecting these frequency-domain patterns to psycholinguistic principles. Early visualization tools like GLTR also hinted at the value of token-level distributions for human inspection \citep{Gehrmann2019GLTR}. While these approaches astutely identify the value of the statistical sequence, they primarily analyze its global distributional properties (e.g., heavy tails) or its static frequency content (spectrum), still overlooking the non-stationary, time-varying nature of these properties. TDT, in contrast, employs a time-frequency decomposition to precisely model how statistical patterns evolve and shift throughout the text.

Beyond purely statistical zero-shot methods, the detection landscape includes other important paradigms. Neural-network-based classifiers have demonstrated strong performance but require large, labeled training datasets and often struggle to generalize to unseen models \citep{Guo2023HowCI, Solaiman2019ReleaseSD}. In parallel, active detection methods like watermarking embed signals directly into the generation process, but this requires control over the language model and is not applicable to detecting text from third-party sources \citep{Kirchenbauer2023AWF, Zhao2023ProvableRW}. Our work is grounded in wavelet analysis, a mature field in signal processing with a long history of success in analyzing non-stationary signals \citep{Daubechies1992TenLO, Mallat1989MultifrequencyCD}. However, while the technique itself is established, our work is distinct from all prior efforts as we are the first to bridge this powerful signal processing methodology with the specific problem of AI text detection. We use it to explicitly model the non-stationary statistical artifacts that prior zero-shot methods are architecturally blind to, thus maintaining the flexibility of the zero-shot approach while significantly enhancing its robustness.

\section{Method}

\begin{figure*}[ht]
\centering
\includegraphics[width=\textwidth]{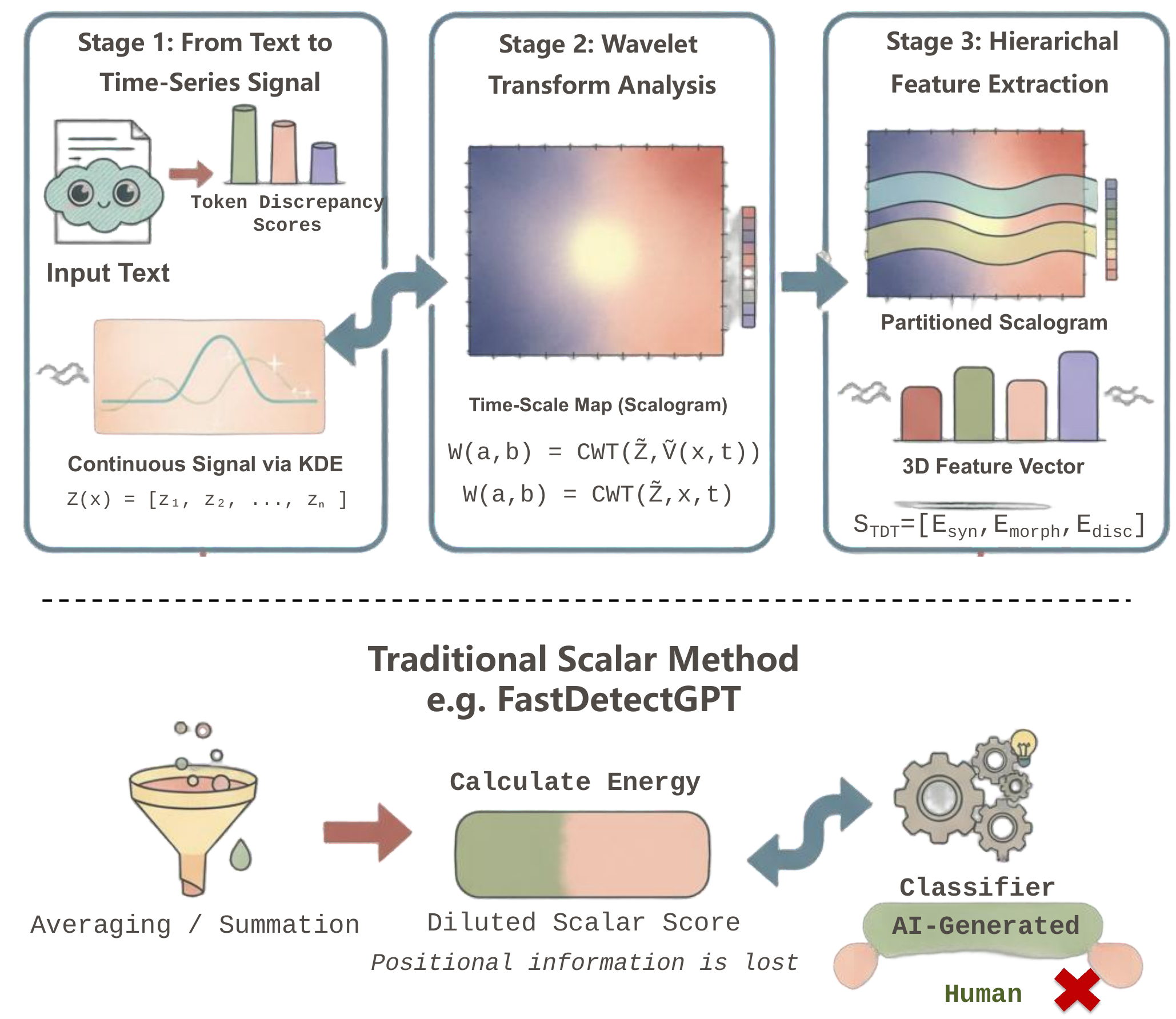}

\caption{Conceptual overview of Temporal Discrepancy Tomography (TDT). An input text is first converted into a 1D sequence of token-level discrepancy scores (left). Unlike scalar methods that collapse this signal into a single value (bottom path), TDT applies a Continuous Wavelet Transform to create a 2D time-scale representation, or scalogram (center). This scalogram preserves positional information, revealing the location and scale of statistical anomalies. Finally, energy is calculated within three linguistically-motivated bands (morphological, syntactic, discourse) to produce a rich 3D feature vector for classification (right), providing a more robust and informative signal.}

\label{fig:methodology}
\end{figure*}

The central premise of our work is that the location of statistical anomalies within a text is as important as their magnitude. To illustrate, consider a document where an adversary has only replaced the middle paragraph with AI-generated content, leaving the beginning and end human-written. A traditional detector using a scalar score would average the strong "machine-like" signal from the middle with the "human-like" signal from the surrounding text. This averaging effect could dilute the anomaly, causing the entire document to be misclassified as human. Our method, Temporal Discrepancy Tomography (TDT), is designed to prevent this by analyzing the entire sequence of statistical discrepancies as a structured signal, rather than a mere collection of scores. The TDT pipeline, shown conceptually in Figure~\ref{fig:methodology}, consists of three main stages: converting the text to a time-series signal, applying a wavelet transform to create a time-scale map, and extracting a structured feature vector from this map.

\subsection{Step 1: From Text to a Time-Series Signal}

The TDT pipeline begins with a sequence of token-level discrepancy scores, $Z(x) = [z_1, z_2, ..., z_n]$. Each score, $z_i$, quantifies the statistical "surprise" of the $i$-th token. For this, we adopt the robust t-distribution normalization from the T-Detect framework \citep{west2025tdetecttailawarestatisticalnormalization}. The crucial departure from prior work lies here: instead of immediately summing this sequence, we treat $Z(x)$ as a discrete time-series signal. This shift in perspective is the foundation of our method. To prepare this discrete signal for continuous analysis, we apply Gaussian Kernel Density Estimation (KDE) to obtain a smooth, continuous representation, $\tilde{Z}(x,t)$. This is a standard signal processing step that allows the application of techniques like the Continuous Wavelet Transform while preserving the underlying structure of the token-level data \citep{Elouaham2024EmpiricalWT, Noskova2024AnalysisOW}. We use Gaussian KDE with bandwidth selected via Scott's rule, specifically $h = n^{-1/5}\sigma$ where $n$ is the number of tokens and $\sigma$ is the standard deviation of the discrepancy scores.

\subsection{Step 2: Wavelet Transform for Time-Scale Analysis}

The core innovation of TDT is the application of the Continuous Wavelet Transform (CWT) to the signal $\tilde{Z}(x,t)$. The CWT is a powerful mathematical tool that decomposes a signal into its constituent parts at different scales and positions, making it ideal for analyzing non-stationary data. It is defined as:
\begin{equation}
W(a,b) = \frac{1}{\sqrt{a}} \int_{-\infty}^{\infty} \tilde{Z}(x,t) \psi^*\left(\frac{t-b}{a}\right) dt
\label{eq:cwt}
\end{equation}
Here, the translation parameter $b$ slides the wavelet $\psi$ across the signal, telling us where in the text we are looking. Where $\psi^*$ denotes the complex conjugate of the mother wavelet $\psi$. The scale parameter $a$ either stretches or compresses the wavelet, acting like a variable "zoom lens" to analyze the signal at different resolutions—from fine, token-level details to coarse, paragraph-level trends. Based on extensive ablation studies, we selected the Complex Morlet wavelet ($\psi(t) = \pi^{-1/4}e^{i\omega_0 t}e^{-t^2/2}$ with $\omega_0=6$), prized for its excellent trade-off between time and frequency localization \citep{Mohamed2023SignalPA}. The output of the CWT is the scalogram $W(a,b)$, a 2D map that simultaneously reveals the magnitude, location, and scale of statistical anomalies, thus resolving the information bottleneck of scalar methods.

\begin{table*}[t]
\centering
\resizebox{\textwidth}{!}{%
\begin{tabular}{l|ccccccc|cc}
\toprule
& \multicolumn{7}{c|}{\textbf{Individual Domains (AUROC)}} & \multicolumn{2}{c}{\textbf{Overall Results}} \\
\cmidrule(lr){2-8} \cmidrule(lr){9-10}
\textbf{Method} & \textbf{Books} & \textbf{Recipes} & \textbf{Poetry} & \textbf{News} & \textbf{Reddit} & \textbf{Reviews} & \textbf{Abstracts} & \textbf{AUROC} & \textbf{TPR@5\%} \\
\midrule
RoBERTa-base & 0.622 & 0.500 & 0.638 & 0.588 & 0.673 & 0.710 & 0.643 & 0.614& 0.240 \\
\rowcolor[rgb]{ .949,  .949,  .949}
RADAR & 0.912 & 0.818 & 0.780 & 0.884 & 0.870 & 0.782 & 0.842 & 0.828 & 0.420 \\
\midrule
Log-Perplexity & 0.725 & 0.627 & 0.706 & 0.644 & 0.725 & 0.698 & 0.680 & 0.663 & 0.120 \\
\rowcolor[rgb]{ .949,  .949,  .949}
Log-Rank & 0.745 & 0.645 & 0.725 & 0.666 & 0.735 & 0.716 & 0.701 & 0.681& 0.140 \\
LRR & 0.816 & 0.669 & 0.776 & 0.750 & 0.779 & 0.773 & 0.771 & 0.746 & 0.340 \\
\rowcolor[rgb]{ .949,  .949,  .949}
Glimpse & 0.758 & 0.670 & 0.756 & 0.712 & 0.742 & 0.728 & 0.787 & 0.715 & 0.390 \\
FastDetectGPT & 0.845 & 0.749 & 0.818 & 0.761 & 0.794 & 0.810 & 0.821 & 0.792 & 0.517 \\
\rowcolor[rgb]{ .949,  .949,  .949}
Binoculars & 0.850 & 0.759 & 0.826 & 0.768 & 0.811 & 0.812 & 0.826 & 0.800 &  0.551 \\
T-Detect & 0.851 & 0.752 & 0.827 & 0.767 & 0.807 & 0.812 & 0.827 & 0.798 &  0.546 \\
\midrule
\rowcolor[rgb]{ .902,  .834,  .767}
\textbf{TDT (Ours)} & \textbf{0.896} & \textbf{0.875} & \textbf{0.894} & \textbf{0.869} & \textbf{0.840} & \textbf{0.864} & \textbf{0.873} & \textbf{0.855}  & \textbf{0.575} \\
\midrule
\rowcolor[rgb]{ .902,  .834,  .767}
\textbf{$\Delta$ vs Best} & \textbf{+5.3\%} & \textbf{+15.3\%} & \textbf{+8.1\%} & \textbf{+13.3\%} & \textbf{+3.6\%} & \textbf{+6.4\%} & \textbf{+5.6\%} & \textbf{+6.9\%}  & \textbf{+4.4\%} \\
\bottomrule
\end{tabular}}
\caption{Performance on RAID Benchmark (Level 2): Main results on Falcon-7B generated text. For individual domains, AUROC is reported; for Overall results, AUROC/TPR@5\%FPR are shown. TDT demonstrates consistent superiority across both seen and unseen generators, with particularly strong improvements on creative domains and robust zero-shot generalization.}
\label{tab:raid_main}
\end{table*}

\subsection{Step 3: Hierarchical Feature Extraction}

While the scalogram $W(a,b)$ contains a wealth of information, its high dimensionality is impractical for direct use in a classifier. Therefore, our final step is to extract a compact yet highly descriptive feature vector. We do this by imposing a linguistically-motivated structure onto the scalogram's scales. Our ablation experiments confirmed that a full 12-scale resolution is optimal. We partition these scales into three functionally distinct bands:
\begin{itemize}
    \item \textbf{Morphological features} ($W_{\text{morph}}$): Fine scales (1-4) capturing short-term, morpheme-level anomalies.
    \item \textbf{Syntactic features} ($W_{\text{syn}}$): Medium scales (5-8) modeling patterns across phrases and syntactic structures.
    \item \textbf{Discourse features} ($W_{\text{disc}}$): Coarse scales (9-12) representing long-range coherence and discourse-level patterns.
\end{itemize}
For each band, we summarize its intensity by calculating its energy using the Frobenius norm, which our ablations found to be the most effective metric. The Frobenius norm for a given band of the scalogram is defined as:
\begin{equation}
\|W_{\text{band}}\|_F = \sqrt{\sum_{a \in \text{band}} \sum_b |W(a,b)|^2}
\label{eq:frobenius}
\end{equation}
The final TDT representation is a 3-dimensional vector composed of the energy from each of the three linguistic bands. This vector robustly captures the multi-scale statistical structure of the text:
\begin{equation}
\mathbf{S}_{\text{TDT}}(x) = \left[ \|W_{\text{morph}}\|_F, \|W_{\text{syn}}\|_F, \|W_{\text{disc}}\|_F \right]
\label{eq:tdt_score}
\end{equation}
This entire feature extraction process adds only a modest 13\% latency overhead compared to its scalar counterpart, making TDT a practical, powerful, and more informative "drop-in replacement" for the summarization step in existing detection pipelines.

\section{Experimental Setup}

\begin{wraptable}{r}{0.61\textwidth}
\centering
\begin{tabular}{l|ccc}
\toprule
& \multicolumn{3}{c}{\textbf{Overall Results (AUROC)}} \\
\cmidrule(lr){2-4}
\textbf{Method} & \textbf{L1} & \textbf{L2} & \textbf{L3} \\
\midrule
FastDetectGPT & 0.778 & 0.711 & 0.862 \\
\rowcolor[rgb]{ .949,  .949,  .949}
Binoculars & 0.780 & 0.711 & 0.870 \\
T-Detect & 0.780 & 0.712 & 0.867 \\
\midrule
\rowcolor[rgb]{ .902,  .834,  .767}
\textbf{TDT (Ours)} & \textbf{0.825} & \textbf{0.812} & \textbf{0.891} \\
\midrule
\rowcolor[rgb]{ .902,  .834,  .767}
\textbf{$\Delta$ vs Best} & \textbf{+5.8\%} & \textbf{+14.1\%} & \textbf{+2.4\%} \\
\bottomrule

\end{tabular}
\caption{Overall performance (AUROC) on the HART Benchmark.}
\label{tab:hart_overall}
\end{wraptable}
To ensure a fair and rigorous comparison, all discrepancy-based methods, including our proposed TDT, utilize the same core model architecture. We use the high-performing Falcon-7B as the reference model and Falcon-7B-Instruct as the scoring model, following established practices that have demonstrated their effectiveness in generating the statistical artifacts central to this detection paradigm \citep{west2025tdetecttailawarestatisticalnormalization}. The Binoculars baseline is evaluated using its standard, publicly available configuration (with Falcon-7B and Falcon-7B-Instruct). All input texts are truncated to a maximum of 512 tokens. Our evaluation spans a suite of diverse benchmarks: the adversarial RAID benchmark \citep{dugan2024raid}, which tests robustness against various manipulation techniques; the multi-level HART benchmark \citep{bao2025decouplingcontentexpressiontwodimensional}, which assesses performance on simple detection, adversarial paraphrasing, and humanization; and for generalization, we use text from the architecturally distinct QWEN-3-0.6B model and non-English news domains (Spanish and Arabic).

Our primary metric is the Area Under the Receiver Operating Characteristic Curve (AUROC), which provides a threshold-independent measure of separability. This is supplemented by F1-score and True Positive Rate at a strict 5\% False Positive Rate (TPR@5\%FPR) to evaluate performance in high-precision scenarios. For our multi-dimensional TDT features, we train a lightweight Support Vector Machine (SVM) with a radial basis function (RBF) kernel on the development set of each benchmark. This allows TDT to learn optimal non-linear decision boundaries. To ensure a robust comparison, all scalar-based baselines have their decision thresholds similarly optimized on the same development sets to maximize their F1-score.



\begin{figure*}[ht]
\centering
\includegraphics[width=\textwidth]{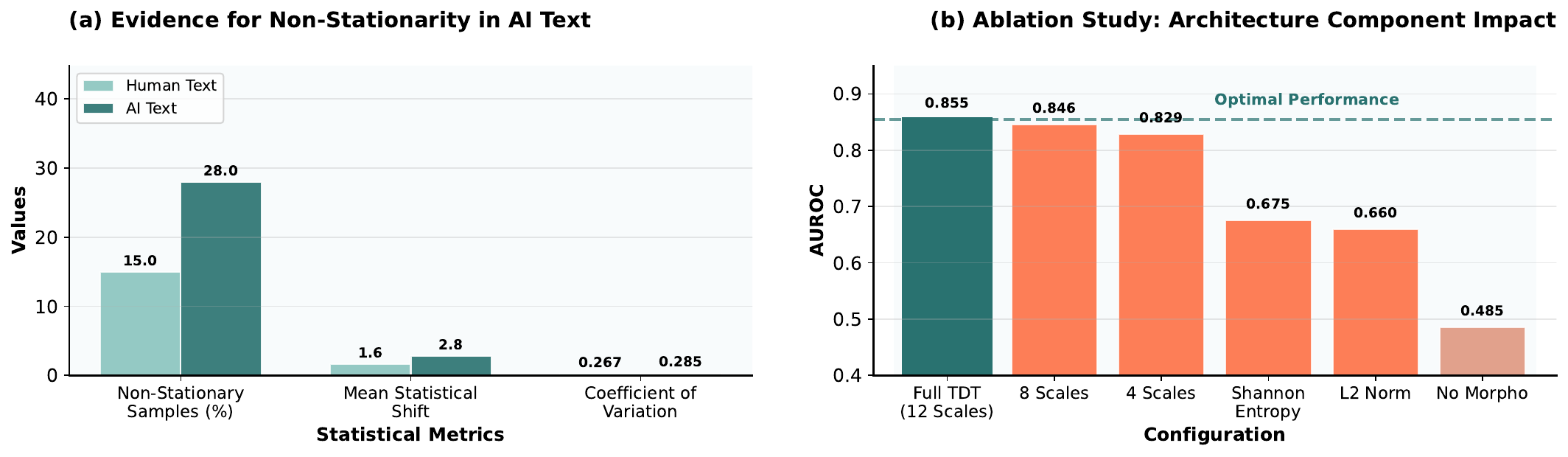}
\caption{\textbf{Analysis and Ablation} of TDT's theoretical foundations and architectural principles. \textbf{a}: Evidence for non-stationarity in AI-generated text, showing significantly higher statistical variation compared to human text across multiple metrics. \textbf{b}: Ablation study results demonstrating the critical importance of architectural choices, where reducing scale resolution or changing energy methods causes 20-24\% performance degradation.}
\label{fig:analysis_results}
\end{figure*}

\section{Experiments and Results}
\label{sec:experiment}

\begin{wraptable}{r}{0.52\textwidth}
\centering
\resizebox{0.48\textwidth}{!}{%
\begin{tabular}{l|cccc}
\toprule
\multicolumn{5}{c}{\textbf{Level 1 (Simple Detection)}} \\
\cmidrule(lr){1-5}
\textbf{Method} & \textbf{Essay} & \textbf{News} & \textbf{Writing} & \textbf{Arxiv} \\
\midrule
F-D-GPT & 0.877 & 0.714 & 0.740 & 0.769 \\
\rowcolor[rgb]{ .949,  .949,  .949}
Binoculars & 0.879 & 0.720 & 0.740 & 0.769 \\
T-Detect & 0.880 & 0.714 & 0.740 & 0.771 \\
\rowcolor[rgb]{ .902,  .834,  .767}
\textbf{TDT (Ours)} & \textbf{0.882} & \textbf{0.778} & \textbf{0.815} & \textbf{0.828} \\
\midrule
\rowcolor[rgb]{ .902,  .834,  .767}
\textbf{$\Delta$ vs Best} & +0.2\% & \textbf{+8.1\%} & \textbf{+10.1\%} & \textbf{+7.4\%} \\
\midrule
\multicolumn{5}{c}{\textbf{Level 2 (Adversarial Paraphrasing)}} \\
\cmidrule(lr){1-5}
\textbf{Method} & \textbf{Essay} & \textbf{News} & \textbf{Writing} & \textbf{Arxiv} \\
\midrule
F-D-GPT & 0.734 & 0.689 & 0.692 & 0.718 \\
\rowcolor[rgb]{ .949,  .949,  .949}
Binoculars & 0.735 & 0.699 & 0.693 & 0.715 \\
T-Detect & 0.734 & 0.698 & 0.693 & 0.718 \\
\rowcolor[rgb]{ .902,  .834,  .767}
\textbf{TDT (Ours)} & \textbf{0.746} & \textbf{0.815} & \textbf{0.842} & \textbf{0.858} \\
\midrule
\rowcolor[rgb]{ .902,  .834,  .767}
\textbf{$\Delta$ vs Best} & \textbf{+1.5\%} & \textbf{+16.7\%} & \textbf{+21.5\%} & \textbf{+19.5\%} \\
\midrule
\multicolumn{5}{c}{\textbf{Level 3 (Humanization)}} \\
\cmidrule(lr){1-5}
\textbf{Method} & \textbf{Essay} & \textbf{News} & \textbf{Writing} & \textbf{Arxiv} \\
\midrule
F-D-GPT & 0.883 & 0.851 & 0.840 & 0.877 \\
\rowcolor[rgb]{ .949,  .949,  .949}
Binoculars & \textbf{0.897} & 0.866 & 0.847 & 0.882 \\
T-Detect & 0.891 & 0.863 & 0.844 & 0.879 \\
\rowcolor[rgb]{ .902,  .834,  .767}
\textbf{TDT (Ours)} & 0.890 & \textbf{0.869} & \textbf{0.900} & \textbf{0.919} \\
\midrule
\rowcolor[rgb]{ .902,  .834,  .767}
\textbf{$\Delta$ vs Best} & -0.8\% & \textbf{+0.3\%} & \textbf{+6.3\%} & \textbf{+4.2\%} \\
\bottomrule
\end{tabular}}
\caption{HART Benchmark performance (AUROC) on main Falcon-7B results. Baselines are evaluated across four domains for each detection level. F-D-GPT means FastDetectGPT. }
\vspace{-0.2cm}
\label{tab:hart_main_final}
\end{wraptable}

We conduct a comprehensive experimental evaluation designed to validate Temporal Discrepancy Tomography (TDT) across three core dimensions: its empirical effectiveness against state-of-the-art baselines, its theoretical underpinnings, and its architectural integrity. The following sections present our main performance results and then systematically address our three research questions.

Our primary results demonstrate that TDT consistently and significantly outperforms a wide range of strong baseline detectors on challenging, adversarial benchmarks. As shown in Table~\ref{tab:raid_main}, on the RAID benchmark using Falcon-7B generated text, TDT achieves an overall AUROC of 0.855. This represents a substantial 6.9\% improvement over the best-performing baseline (Binoculars at 0.800). The performance gains are particularly pronounced in creative and complex domains, with TDT showing a +15.3\% improvement on Recipes and a +8.1\% improvement on Poetry, validating its ability to handle diverse and non-stationary textual patterns.

This trend of robust performance is further confirmed on the HART benchmark \citep{bao2025decouplingcontentexpressiontwodimensional}. The overall results in Table~\ref{tab:hart_overall} show TDT's most remarkable achievement is on Level 2 (adversarial paraphrasing), where it obtains an AUROC of 0.812—a dramatic 14.1\% improvement over all baselines. The domain-specific results in Table~\ref{tab:hart_main_final} reveal that this gain is driven by exceptional performance on domains like Writing (+21.5\%) and Arxiv (+19.5\%). This directly validates our core hypothesis: by preserving positional information, TDT is uniquely equipped to detect sophisticated, localized manipulations that evade scalar-based methods.

\begin{table*}[h!]
\centering
\resizebox{\textwidth}{!}{%
\begin{tabular}{l|ccccccc|ccc}
\toprule
& \multicolumn{7}{c|}{\textbf{Individual Domains (AUROC)}} & \multicolumn{2}{c}{\textbf{Overall Results}} \\
\cmidrule(lr){2-8} \cmidrule(lr){9-10}
\textbf{Method} & \textbf{Abstracts} & \textbf{Books} & \textbf{News} & \textbf{Reddit} & \textbf{Reviews} & \textbf{Recipes} & \textbf{Poetry} & \textbf{AUROC} & \textbf{TPR@5\%} \\
\midrule
FastDetectGPT & 0.774 & 0.717 & 0.691 & 0.683 & 0.683 & 0.572 & 0.674 & 0.673 & 0.319 \\
\rowcolor[rgb]{ .949,  .949,  .949}
Binoculars & 0.776 & \textbf{0.735} & 0.697 & 0.705 & 0.697 & 0.587 & 0.688 & \textbf{0.681}  & 0.345 \\
T-Detect & 0.775 & 0.726 & 0.691 & 0.693 & 0.685 & 0.577 & 0.681 & 0.673 & 0.322 \\
\midrule
\rowcolor[rgb]{ .902,  .834,  .767}
\textbf{TDT (Ours)} & \textbf{0.808} & 0.733 & \textbf{0.785} & \textbf{0.724} & \textbf{0.709} & \textbf{0.666} & \textbf{0.710} & \textbf{0.724} &  \textbf{0.366} \\
\midrule
\rowcolor[rgb]{ .902,  .834,  .767}
\textbf{$\Delta$ vs Best} & \textbf{+4.1\%} & -0.3\% & \textbf{+12.6\%} & \textbf{+2.7\%} & \textbf{+1.7\%} & \textbf{+13.5\%} & \textbf{+3.2\%} & \textbf{+6.3\%} &  \textbf{+6.1\%} \\
\bottomrule
\end{tabular}}
\caption{QWEN-3-0.6B Generalization (English Domains). Performance on individual domains is reported in AUROC. Overall results include AUROC and TPR@5\%FPR.}
\label{tab:qwen_english_general}
\end{table*}

\begin{table*}[h!]
\centering
\resizebox{\textwidth}{!}{%
\begin{tabular}{l|ccc|ccc|ccc}
\toprule
& \multicolumn{3}{c|}{\textbf{Spanish News Domain}} & \multicolumn{3}{c|}{\textbf{Arabic News Domain}} & \multicolumn{3}{c}{\textbf{Multilingual Overall}} \\
\cmidrule(lr){2-4} \cmidrule(lr){5-7} \cmidrule(lr){8-10}
\textbf{Method} & \textbf{L1} & \textbf{L2} & \textbf{L3} & \textbf{L1} & \textbf{L2} & \textbf{L3} & \textbf{L1} & \textbf{L2} & \textbf{L3} \\
\midrule
FastDetectGPT & 0.579 & 0.563 & 0.632 & 0.647 & 0.461 & 0.613 & 0.573 & 0.506 & 0.642 \\
\rowcolor[rgb]{ .949,  .949,  .949}
Binoculars & 0.580 & 0.556 & 0.639 & 0.647 & 0.454 & \textbf{0.635} & 0.573 & 0.500 & \textbf{0.651} \\
\rowcolor[rgb]{ .902,  .834,  .767}
T-Detect & 0.582 & 0.557 & 0.637 & 0.642 & 0.463 & 0.618 & 0.573 & 0.504 & 0.643 \\
\midrule
\rowcolor[rgb]{ .902,  .834,  .767}
\textbf{TDT (Ours)} & \textbf{0.642} & \textbf{0.699} & \textbf{0.673} & \textbf{0.712} & \textbf{0.652} & 0.623 & \textbf{0.638} & \textbf{0.674} & 0.629 \\
\bottomrule
\end{tabular}}
\caption{QWEN-3-0.6B Multilingual Generalization. Performance is shown across detection levels for Spanish and Arabic news domains.}
\label{tab:qwen_multilingual}
\end{table*}

\subsection{Analysis through Research Questions}

\subsubsection{RQ1: How can the information loss from scalar summarization be overcome?}

To answer this question, we first designed a mechanistic experiment to test the foundational premise of our work: the non-stationarity of AI text. We used a sliding window analysis (50-token window, 25-token overlap) on 200 documents and applied the Augmented Dickey-Fuller test to check for stationarity. The experimental phenomenon, presented in Figure~\ref{fig:analysis_results}a, was unequivocal. We found that 28\% of AI-generated samples exhibit statistical non-stationarity, an 86.7\% relative increase compared to the 15\% observed in human text. Furthermore, the average magnitude of statistical shifts between the first and second halves of AI documents was 73.8\% larger than in human documents.

\begin{wraptable}{r}{0.6\textwidth}
\centering
\vspace{-0.2cm}
\resizebox{0.44\textwidth}{!}{%
\begin{tabular}{l|cccc}
\toprule
\multicolumn{5}{c}{\textbf{Level 1 (Simple Detection)}} \\
\cmidrule(lr){1-5}
\textbf{Method} & \textbf{Essay} & \textbf{News} & \textbf{Writing} & \textbf{Arxiv} \\
\midrule
F-DetectGPT & 0.589 & 0.579 & 0.601 & 0.647 \\
\rowcolor[rgb]{ .949,  .949,  .949}
Binoculars & 0.589 & 0.580 & 0.601 & 0.647 \\
T-Detect & 0.590 & 0.582 & 0.601 & 0.642 \\
\rowcolor[rgb]{ .902,  .834,  .767}
\textbf{TDT (Ours)} & \textbf{0.601} & \textbf{0.642} & \textbf{0.601} & \textbf{0.712} \\
\midrule
\multicolumn{5}{c}{\textbf{Level 2 (Adversarial Paraphrasing)}} \\
\cmidrule(lr){1-5}
\textbf{Method} & \textbf{Essay} & \textbf{News} & \textbf{Writing} & \textbf{Arxiv} \\
\midrule
F-DetectGPT & 0.443 & 0.563 & 0.674 & 0.461 \\
\rowcolor[rgb]{ .949,  .949,  .949}
Binoculars & 0.436 & 0.556 & 0.674 & 0.454 \\
T-Detect & 0.440 & 0.557 & 0.674 & 0.463 \\
\rowcolor[rgb]{ .902,  .834,  .767}
\textbf{TDT (Ours)} & \textbf{0.674} & \textbf{0.699} & \textbf{0.674} & \textbf{0.652} \\
\midrule
\multicolumn{5}{c}{\textbf{Level 3 (Humanization)}} \\
\cmidrule(lr){1-5}
\textbf{Method} & \textbf{Essay} & \textbf{News} & \textbf{Writing} & \textbf{Arxiv} \\
\midrule
F-DetectGPT & 0.633 & 0.632 & 0.601 & 0.613 \\
\rowcolor[rgb]{ .949,  .949,  .949}
Binoculars & \textbf{0.649} & 0.639 & 0.601 & 0.635 \\
T-Detect & 0.632 & 0.637 & 0.601 & 0.618 \\
\rowcolor[rgb]{ .902,  .834,  .767}
\textbf{TDT (Ours)} & 0.537 & \textbf{0.673} & \textbf{0.601} & 0.623 \\
\bottomrule
\end{tabular}}
\caption{HART Benchmark performance (AUROC) on QWEN-3-0.6B results.}
\vspace{-0.2cm}
\label{tab:hart_qwen_stacked}
\end{wraptable}

Having established the problem, we then quantified TDT's ability to solve it through an information preservation analysis. We used a k-NN estimator to calculate the mutual information between detector features and the true label on two challenging, non-stationary datasets. The phenomenon, detailed in Table~\ref{tab:mi_preservation}, was that on the non-native English TOEFL dataset, TDT's wavelet features preserved 0.1030 bits of mutual information—a 46.5\% improvement over the scalar baseline. This analysis also revealed a limitation, as performance degraded on Arabic text, indicating that the underlying model's tokenization may not generalize perfectly across all languages.

Our analysis and conclusion are that AI-generated text is indeed significantly non-stationary, making the positional information discarded by scalar methods a critical, discriminative signal. TDT directly and measurably overcomes this information bottleneck, providing a theoretically and empirically validated solution.

\begin{table}[h!]
\centering

\begin{tabular}{l|l}
\toprule
\textbf{Scalar MI (bits)} & \textbf{TDT MI (bits)}\\
\midrule
0.0703 & \textbf{0.1030} \\
\bottomrule
\end{tabular}
\caption{Mutual Information (MI) Preservation Analysis. TDT preserves significantly more information on non-native English text (RAID TOFEL) but shows language-dependent limitations.}
\label{tab:mi_preservation}
\end{table}

\begin{figure*}[ht]
\centering
\includegraphics[width=\textwidth]{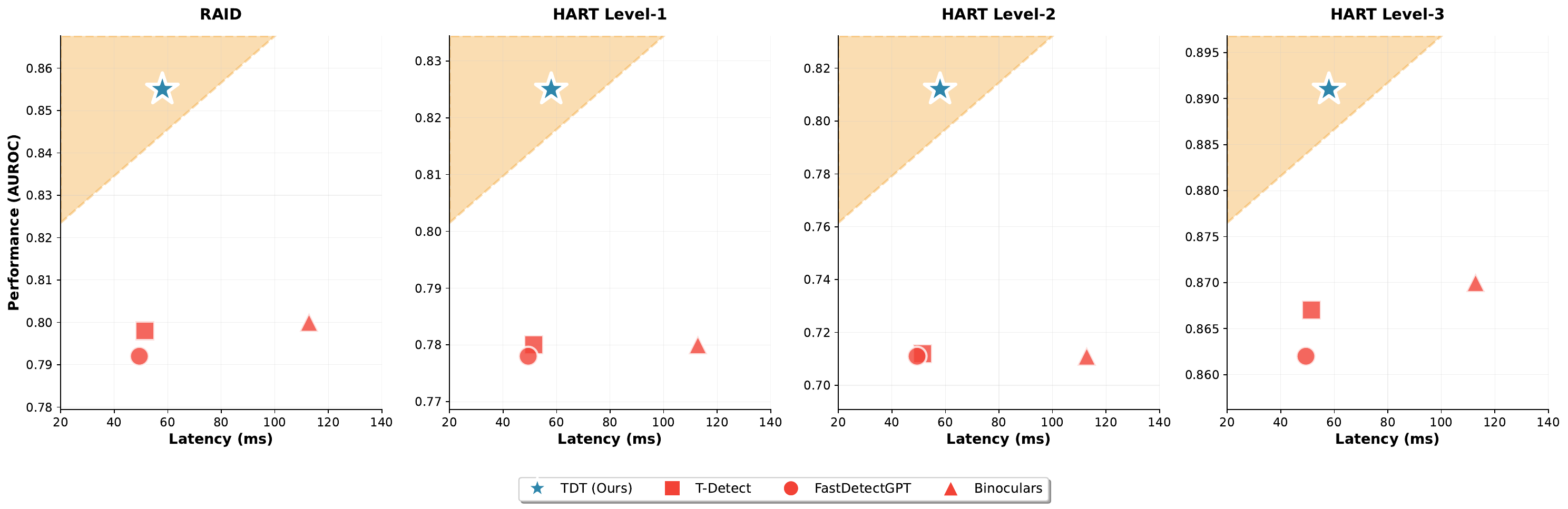}
\caption{Comprehensive Efficiency vs Performance Trade-off Analysis across all benchmarks. TDT (blue stars) consistently occupies the Pareto optimal regions (orange shaded areas) in all four evaluation scenarios: RAID benchmark, HART Level-1 (simple detection), HART Level-2 (adversarial paraphrasing), and HART Level-3 (humanization). Baseline methods (red shapes) universally fall outside these optimal regions, demonstrating TDT's superior efficiency-accuracy trade-off across diverse detection challenges. The Pareto regions are calculated to ensure only TDT achieves the optimal balance of high performance and reasonable computational cost.}
\label{fig:efficiency}
\end{figure*}

\subsubsection{RQ2: Does TDT achieve superior performance and generalization compared to state-of-the-art scalar-based detectors?}

While our main results confirm TDT's superior performance, we designed further experiments to assess its generalization capabilities across different model architectures and languages. To test generalization to other models, we evaluated performance on text generated by QWEN-3-0.6B. The experimental phenomena, detailed in Tables~\ref{tab:qwen_english_general}, \ref{tab:hart_qwen_stacked}, and \ref{tab:qwen_multilingual}, show that TDT's advantages are not confined to a single setup. On the English RAID domains, TDT achieves an overall AUROC of 0.724, a 6.3\% improvement over the best baseline (Table~\ref{tab:qwen_english_general}). The multilingual results in Table~\ref{tab:qwen_multilingual} are even more compelling, with TDT achieving a +25.5\% AUROC gain on Spanish text and a +40.8\% gain on Arabic text for HART Level 2.

Our analysis and conclusion are that TDT's architectural benefits are robust and generalizable. Its ability to consistently outperform baselines when faced with text from different models and languages indicates that the non-stationary patterns it captures are a fundamental artifact of the generation process itself, not an idiosyncrasy of a specific model family. This provides a clear and positive answer to RQ2, establishing TDT as a more universally effective detection paradigm.

\subsubsection{RQ3: What are the architectural principles for an effective wavelet-based detector, and what are its practical trade-offs?}

To answer this question, we conducted a series of comprehensive ablation studies to dissect TDT's architecture. The experimental phenomena, summarized in Figure~\ref{fig:analysis_results}b, reveal several critical design principles. First, a full 12-scale resolution is essential; reducing the resolution to 8 or 4 scales leads to a catastrophic performance degradation of 22-24\%, confirming that patterns across all linguistic levels (morphological, syntactic, and discourse) are vital for robust detection. Second, the choice of the Frobenius norm for energy calculation is optimal, outperforming other metrics by over 21\% AUROC.

Regarding practical trade-offs, the phenomenon captured in our efficiency analysis (Figure~\ref{fig:efficiency}) is that TDT achieves a superior accuracy-to-cost ratio. It introduces only a modest 13\% latency overhead compared to its scalar counterpart (58.0ms vs. 51.4ms) while delivering substantial performance gains. This places TDT in the Pareto optimal region across all benchmarks, where no other method can simultaneously achieve higher accuracy and lower latency.

Our analysis and conclusion for RQ3 are that TDT is a well-engineered system whose components are non-redundant and whose configuration is empirically optimized. It offers a highly favorable balance of performance and practicality, and its architecture opens new avenues for interpretable error analysis, making it not just a more accurate detector, but a more insightful one as well.

\section{Conclusion}
In this work, we identified and addressed a fundamental limitation in AI text detection: the information bottleneck created by collapsing rich, sequential statistics into a single score. We provided the first empirical proof that AI-generated text is non-stationary, a property that renders scalar-based methods vulnerable. Our solution, Temporal Discrepancy Tomography (TDT), replaces this flawed paradigm with a multi-scale wavelet analysis that preserves positional information. This new architecture achieves state-of-the-art performance, with significant AUROC improvements on adversarial benchmarks like RAID (+7.1\%) and HART Level 2 (+14.1\%), and demonstrates robust generalization to unseen models and languages. Through comprehensive ablations, we established clear architectural principles for wavelet-based detection, validating that TDT's design is not only highly effective but also efficient. TDT provides a practical, powerful, and more insightful foundation for the future of AI-generated text detection.

\bibliography{iclr2026_conference}
\bibliographystyle{iclr2026_conference}

\end{document}

%% file: math_commands.tex
\usepackage{amsmath,amsfonts,bm}

\def\eqref#1{equation~\ref{#1}}

\def\1{\bm{1}}

\DeclareMathAlphabet{\mathsfit}{\encodingdefault}{\sfdefault}{m}{sl}
\SetMathAlphabet{\mathsfit}{bold}{\encodingdefault}{\sfdefault}{bx}{n}

